\documentclass{article}
\pdfoutput=1
\usepackage{amsmath,amssymb,amsfonts}
\usepackage{algorithmic}
\usepackage{graphicx}
\usepackage{textcomp}
\usepackage{xcolor}

\usepackage{spconf,amsmath,graphicx}
\usepackage[square,numbers,sort]{natbib}
\usepackage[obeyspaces]{url}
\usepackage[utf8]{inputenc}
\usepackage{tikz}
\usetikzlibrary{external}

\begin{document}
\title{PhishGAN: Data Augmentation and Identification of Homoglyph Attacks\\}

\name{}
\address{}

\makeatletter
\let\@oldmaketitle\@maketitle
\renewcommand{\@maketitle}{\@oldmaketitle
  \vspace{-2.2cm}
  \includegraphics[width=\linewidth]{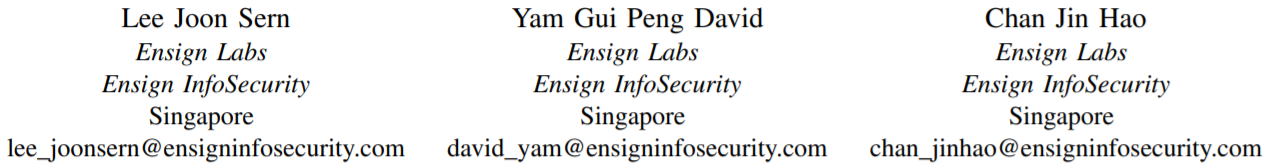}\bigskip}
\makeatother
\maketitle

\begin{abstract}
    Homoglyph attacks are a common technique used by hackers to conduct phishing. Domain names or links that are visually similar to actual ones are created via punycode to obfuscate the attack, making the victim more susceptible to phishing. For example, victims may mistake "|inkedin.com" for "linkedin.com" and in the process, divulge personal details to the fake website. Current State of The Art (SOTA) typically make use of string comparison algorithms (e.g. Levenshtein Distance), which are computationally heavy. One reason for this is the lack of publicly available datasets thus hindering the training of more advanced Machine Learning (ML) models. Furthermore, no one font is able to render all types of punycode correctly, posing a significant challenge to the creation of a dataset that is unbiased toward any particular font. This coupled with the vast number of internet domains pose a  challenge in creating a dataset that can capture all possible variations. Here, we show how a conditional Generative Adversarial Network (GAN), PhishGAN, can be used to generate images of hieroglyphs, conditioned on non-homoglyph input text images. Practical changes to current SOTA were required to facilitate the generation of more varied homoglyph text-based images. We also demonstrate a workflow of how PhishGAN together with a Homoglyph Identifier (HI) model can be used to identify the domain the homoglyph was trying to imitate. Furthermore, we demonstrate how PhishGAN's ability to generate datasets on the fly, facilitating the quick adaptation of cybersecurity systems to detect new threats as they emerge. \footnote{© 2020 IEEE.  Personal use of this material is permitted.  Permission from IEEE must be obtained for all other uses, in any current or future media, including reprinting/republishing this material for advertising or promotional purposes, creating new collective works, for resale or redistribution to servers or lists, or reuse of any copyrighted component of this work in other works.}
\end{abstract}


\section{Introduction}
    A common type of phishing attack involves permuting alphabets of the same Latin characters family. These are also commonly known as look-alike domains, or typo-squatting.
    In a study by \citeauthor{dhamija2006phishing} \cite{dhamija2006phishing}, researchers fooled 90.9\% of their participants by hosting a website at "www.bankofthevvest[.]com", with two "v's instead of a "w" in the domain name, showcasing the effectiveness of this strategy.
    
    For lookalike and typo-squatting attack types, the current industry accepted approach is to calculate the edit distance between strings. Equation \eqref{eq1} shows two sample domains which are 1 Levenshtein edit distance away from "facebook.com". In this method, a lower Levenshtein value indicates more similar domains, increasing the confidence of a phishing attempt.
    \begin{equation}
    \begin{split}
        ld("facebook.com", "face4book.com") = 1\\
        ld("facebook.com", "faceb0ok.com") = 1 \label{eq1}
    \end{split}
    \end{equation}
    
    However, the edit distance method fails when the attackers utilize homoglyph attacks (a subset of lookalike attacks), which uses characters not within the Latin characters group. For example, the domains "fácebook.com" and "facebooZ.com" while both 1 edit distance from "facebook.com", have different visual characteristics.

    As most modern browsers support the display of Internationalized Domain Names (IDN), domain names with digits and other special characters can all be registered. IDNs are converted to their Latin character equivalent in the form of punycodes. Though extremely useful in facilitating domain names of various languages, it opens up the possibility of cyber attacks, particularly, homoglyph attacks. The homoglyph attack vector comes into play when there is a mixture of characters that look similar to their Latin counterparts. As shown in Table \ref{tab:domains_with_replaced}, it is not easy to differentiate the homoglyphs and their original domains.

    \begin{table}[!htbp]
    \caption{Domains with Replaced Characters}
    \label{tab:domains_with_replaced}
    \centerline{
        \begin{tabular}{|p{2cm}|p{1.4cm}|p{2cm}|p{1.9cm}|}
        \hline
        \textbf{Original} & \textbf{Replaced} & \textbf{Punycode} & \textbf{Visualized}\\
        \hline
        facebook.com & "a" to "á" & xn\texttt{-{}-}fcebook-hwa.com & fácebook.com\\
        google.com & "l" to "ł" & xn\texttt{-{}-}googe-n7a.com & googłe.com\\
        imda.gov.sg & "i" to "ı" & xn\texttt{-{}-}mda-iua.gov.sg & ımda.gov.sg\\
        \hline
        \end{tabular}
        }
    \end{table}
    
    As homoglyph attacks have been on the rise since 2000\footnote{https://www.digitalinformationworld.com/2020/03/phishing-hackers-continue-to-use-homoglyph-characters-in-domain-names-to-trick-users.html}, Many techniques have been proposed in the literature to detect such attacks. \citeauthor{shamfinder} \cite{shamfinder} studied the similarity between single characters and evaluated their pair wise similarity based on mean squared error. They then tuned their algorithm by getting humans to evaluate similarity of character pairs. A major drawback of this is that the string of words wasn't taken into account as similarity comparisons was done at character level. Furthermore, the authors opined that a combination of homoglyphs could affect the confusability of homoglyph strings. In this work, the lack of a large dataset was a major problem, hence the need for human labellers.
    
    \citeauthor{woodbridge2018detecting} \cite{woodbridge2018detecting} showed that a Siamese CNN was able to detect and classify homoglyph attacks. He also contributed a dataset containing pairs of real and spoofed domains renderable in Arial. Though extremely useful for the purpose of training ML algorithms, the major drawback is that it is inherently biased towards Arial font. This means that punycode that could be rendered by other fonts are not taken into account and deep learning models trained on such a dataset would have a bias towards Arial font. Also, creating a curated dataset for multiple fonts would be extremely tedious and may not be efficient, as it would again be biased towards those fonts. 
    
    Thus, we propose to make use of state-of-the-art GAN algorithms to extend \citeauthor{woodbridge2018detecting}'s dataset to produce potentially infinite possibilities of homoglyphs. Though there may not be a way to render these GAN generated homoglyphs in punycode, we expect that GAN generated data would provide significantly more variability to the dataset such that ML algorithms would not be constrained or limited to variations of any particular font.
    
    We highlight our 3 main contributions. Our major contribution is PhishGAN, which can generate realistic text images of homoglyphs. We show that PhishGAN is able to produce a more varied set of homoglyph images than naively applying SOTA algorithms like Pix2Pix and CycleGAN. Although there may not be a valid punycode to produce PhishGAN's output images, it is extremely useful in serving the purpose of dataset augmentation for data-intensive deep learning algorithms to train on.
    
    Our second contribution is an extensive validation of PhishGAN's output via a Homoglyph Identifier (HI), which is intended to detect and classify which domain a homoglyph was trying to mimic. We aim to show that a model would be able to identify and classify real-life homoglyphs after being trained on just data generated by PhishGAN. Other than using triplet loss over paired loss, the HI is largely similar to \citeauthor{woodbridge2018detecting}'s work.

    Our final contribution is a realistic scenario testing, showing how PhishGAN's ability to generate dataset on the fly can help cybersecurity systems adapt quickly to new emerging threats.
\section{Preliminaries and Problem Setup}
    In this section, we briefly review the current SOTA conditional GANs, particularly Pix2Pix and CycleGAN and introduce the mathematical formulation of our problem.
    
    Pix2Pix and CycleGAN are the closest related works to PhishGAN and they belong to the family of conditional GANS, whose outputs are conditioned on an input. These 2 models were experimented on extensively to determine the vital components to produce visually realistic homoglyphs.
    
\begin{figure}[!htp]
        \centering
        \includegraphics[width=7cm]{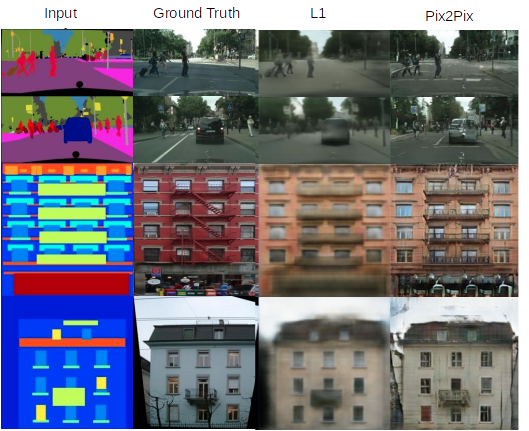}
        \caption{Pix2Pix generating realistic images from sketches}
        \label{fig:pix2pix_sample}
    \end{figure}

    Pix2Pix is a conditional GAN algorithm developed by \citeauthor{isola2017image} \cite{isola2017image} with the aim to translate input images into realistic output images. For example, given outline sketches of objects, it is able to include colours and produce realistic looking images as shown in Figure \ref{fig:pix2pix_sample}. \citeauthor{isola2017image} made use of paired images; his dataset contains input sketch-like images and desired photo realistic ground truth images. As the images are paired, one could simply minimise the $L1$ loss between the ground truth and the network output described in \eqref{eq2}

    \begin{figure}[!htp]
        \centering
        \includegraphics[width=8cm]{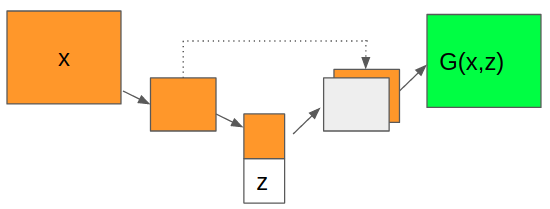}
        \caption{UNet Neural Network Architecture}
        \label{fig:unet_architecture}
    \end{figure}
    
    \begin{equation}
        L_{L1} = || y-G(x,z) || \label{eq2}
    \end{equation}
    
    In \eqref{eq2}, $x$ is the input sketch like image, $z$ is the noise vector added to allow for variation, $G$ is the generator function that generates an output image given $x$ and $z$ as inputs. Finally, $y$ is the target output image. The generator function, $G$, is typically realised via a UNet CNN architecture as shown in Figure  \ref{fig:unet_architecture}. An input image $x$ is first passed through CNN layers to reduce its dimensions to a low dimensional space. At the smallest dimension, a noise tensor is concatenated channel-wise and it is then upsampled via convolution transpose layers to produce a tensor of the same size and shape as the input $x$. At each upsampling stage, the corresponding tensor at the downsampling stage is concatenated channel-wise to the upsampled tensor. Like ResNet architectures \cite{resnet_ref}, UNets facilitate gradient backpropagation through neural networks.
    
    \citeauthor{isola2017image} showed that with the above architecture and simply minimising the $L1$ loss function, blurry images could be produced (see Figure \ref{fig:pix2pix_sample}). \citeauthor{isola2017image} then added a discriminator, trained to classify whether images are generated or real. This is easily achieved by training a CNN in a supervised way (i.e. With a discriminator function, $D$, that tags "real" images as "0" and "fake" images generated by the generator as "1"). The generator's objective function is then augmented with a loss that describes its ability to fool the discriminator in \eqref{eq3}.
    \begin{equation}
        G^* = min_G log(D(x,y)) + log(1-D(G(x,z))) + L_{L1} \label{eq3}
    \end{equation}
    
    The discriminator objective function\footnote{Note that the discriminator function also takes in the conditional image $x$, indicating that its classification is also conditioned on $x$.} is as follows:
    \begin{equation}
        D^* = max_D log(D(x,y)) + log(1-D(G(x,z))) \label{eq4}
    \end{equation}

\begin{figure}[!htp]
        \centering
        \includegraphics[width=3.5cm]{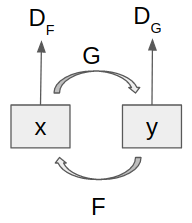}
        \caption{High level training concept for CycleGAN}
        \label{fig:cyclegan}
    \end{figure}

    \citeauthor{zhu2017unpaired} \cite{zhu2017unpaired} relaxed the need for paired images in CycleGAN; it only requires samples of $x$'s and $y$'s. It is essentially a combination of 2 Pix2Pix networks as shown in Figure \ref{fig:cyclegan}. In particular $G$ and $D_G$ are as previously described by Pix2Pix where $G$ aims to generate $y$, with input $x$, while $D_G$ aims to determine whether the $y$'s are generated by $G$ or not. There is however, another set of generator ($F$) and discriminator ($D_F$). The function, $F$, aims is to generate $x$ from $y$ and $D_F$ aims to determine if $x$'s are generated by $F$ or not. 

    The objective function for the discriminators are same as before, while the generators have 2 loss terms in place of $L1$:
    \begin{enumerate}
        \item Cycle Consistency loss: $L_c=||F(G(x,z_G))-x|| + ||G(F(y,z_F))-y||$. This loss\footnote{$z_G$ and $z_F$ refer to the noise tensors added to generator functions G and F respectively} puts a constraint such that an input image $x$ after being processed by $G$ and $F$ sequentially should be $x$. Similarly, an input image $y$, after being processed by $F$ and $G$ should be $y$. This loss acts as a regularisation loss to limit the search space while optimizing $G$ and $F$. It is added to the loss function of both $G$ and $F$.
        \item Identity losses: $L_{I,G}=||G(y,z_G)-y||, L_{I,F}= ||F(x,z_F)-x||$. These constrains that passing an image $y$ to $G$ should yield y as $G$ aims to produce images that are from the $y$'s dataset, and vice versa for $F$. $L_{I,G}$ is added to the loss function of $G$ while $L_{I,F}$ is added to the loss function of $F$.
    \end{enumerate}

    \citeauthor{zhu2017unpaired} \cite{zhu2017unpaired} showed that CycleGAN was able to "morph" images of horses ($x$) to zebras ($y$), without paired images.
    
    Both Pix2Pix and CycleGANs have been tested on photo images. However, to the best of our knowledge, there is no GAN in the literature that looks at morphing text based images. In this work, we investigate the applicability of both Pix2Pix and CycleGAN for our use case of generating "glyphed" text images from real images. The ability to do this will essentially allow us to generate an infinite sized homoglyph dataset.
    
    Related to our second and third contributions is \citeauthor{woodbridge2018detecting}'s \cite{woodbridge2018detecting} work on using Siamese CNN to detect homoglyph strings by checking whether the $L2$ distance of their encodings versus the encodings of a checking list of strings is below a certain threshold\footnote{If so, a homoglyph detection is flagged and it is classified as the checking list item with the smallest $L2$ distance.}.  Siamese CNN aims to extract features from images into a single vector. This is represented as follows: $S(i) = e$ where $S$ is the CNN function that encodes an image $i$ into encoding $e$. The network $S$ is then trained on pair loss, which is the $L2$ distance between 2 encoding. For pairs of strings that have been labelled as similar, we minimise the $L2$ distance; while for pairs of strings that are dissimilar, we maximise the $L2$ distance. This loss function is also known as contrastive loss. A drawback is that it doesn"t place an upper bound on how far to segregate 2 differing points. Thus, it isn"t optimal if 2 points which are already at the centre of their respective clusters are pushed further away and out of their clusters. The triplet loss has been shown to produce better encoding than the contrastive loss over a variety of use cases \cite{triple_loss_v_contrastive_loss}. Thus, in this work, we make use of the triplet loss to train  $S$\footnote{As mentioned earlier, we will not be comparing the advantages of the triplet loss over the contrastive loss in this work. Instead, we will use it to train a HI to show possible use cases of PhishGAN and also to validate its output.}. 

\section{Methodology}
    \subsection{PhishGAN}
\begin{figure}[!htp]
        \centering
        \includegraphics[width=9cm]{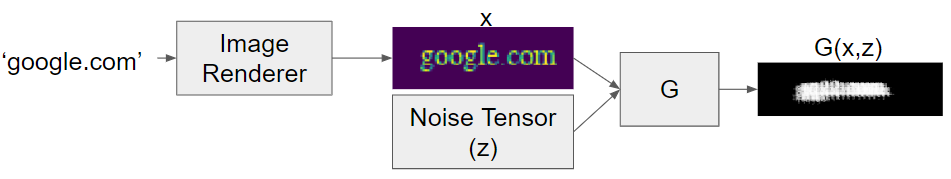}
        \caption{Workflow for PhishGAN}
        \label{fig:PhishGAN_workflow}
    \end{figure}

    PhishGAN aims to generate homoglyphs given any Latin based input text string. To overcome the bias a manually created dataset may have toward any one font, PhishGAN should also be able to accept strings of multiple fonts and output homoglyphs corresponding to such fonts. We make use of \citeauthor{woodbridge2018detecting}'s contributed domain dataset that contains pairs of domains and possible homoglyphs renderable by Arial font. Our workflow for PhishGAN is shown in Figure \ref{fig:PhishGAN_workflow}.
    
     A domain string is first rendered into a greyscale image (of shape 40 $\times$ 400 $\times$ 1) using a certain font via the Python Pillow package. We also perform data augmentation by randomly shifting the text around the 40 $\times$ 400 $\times$ 1 image. Next, the rendered image, $x$, and a noise tensor ($z$) is fed to a generator function (which uses the UNet architecture) to produce image $G(x,z)$. Our UNet architecture is shown in Table \ref{tab:unet_architecture}. Batch normalisation is done between every convolutional layer.
    
    \begin{table}[!htbp]
    \caption{UNet Architecture}
    \label{tab:unet_architecture}
    \centerline{
        \begin{tabular}{|p{0.9cm}|p{0.9cm}|p{1cm}|p{2.5cm}|p{1.5cm}|}
        \hline
        \textbf{Filters} & \textbf{Stride} & \textbf{Kernel} & \textbf{Convolution Type} & \textbf{Dropout Rate}\\
        \hline
        16 & [1,2] & [10,10] & Conv2D & 0 \\
        16 & [2,2] & [10,10] & Conv2D & 0 \\
        16 & [2,2] & [10,10] & Conv2D & 0 \\
        16 & [2,2] & [10,10] & Conv2D & 0 \\
        256 & [2,2] & [10,10] & Conv2DTranspose & 0.1 \\
        128 & [2,2] & [10,10] & Conv2DTranspose & 0.1 \\
        64 & [2,2] & [10,10] & Conv2DTranspose & 0.1 \\
        1 & [1,2] & [10,10] & Conv2DTranspose & 0.1 \\
        \hline
        \end{tabular}
        }
    \end{table}

    The input to the UNet is the 40 $\times$ 400 $\times$ 1 greyscale image, normalized to between $-1$ and $1$. 512 channels of randomly generated Gaussian noise were concatenated channel-wise to the tensor output of the final 2D Convolution layer. As such 528 channels will be fed to the first convolution transpose layer to eventually reconstruct a 40 $\times$ 400 $\times$ 1 image tensor. Leaky Relu was used as the activation function between each layer, except the final layer where a Tanh activation was used.
 
     \begin{figure}[!htp]
        \centering
        \includegraphics[width=7cm]{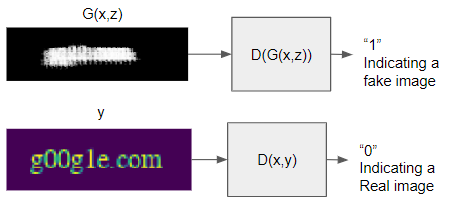}
        \caption{Sample Real \& Generated images}
        \label{fig:discriminator_example}
    \end{figure}
    
    Next, a discriminator was trained to identify whether the input image is real or fake as shown in Figure \ref{fig:discriminator_example}. Table \ref{tab:discriminator_architecture} shows the details of the discriminator network architecture. Again, the Leaky Relu activation function was used for all layers except the final layer, which was activated via a Sigmoid function, no drop out was used. We also find that addition of batch normalisation here significantly degrades final performance. As such, instance normalisation was used instead.
    
    \begin{table}[!htbp]
    \caption{Discriminator Architecture}
    \label{tab:discriminator_architecture}
    \centerline{
        \begin{tabular}{|p{0.9cm}|p{0.9cm}|p{1cm}|p{2cm}|}
        \hline
        \textbf{Filters} & \textbf{Stride} & \textbf{Kernel} & \textbf{Convolution Type}\\
        \hline
        16 & [1,2] & [3,5] & Conv2D \\
        32 & [1,2] & [3,5] & Conv2D \\
        32 & [2,2] & [3,5] & Conv2D \\
        64 & [2,2] & [3,5] & Conv2D \\
        64 & [1,2] & [3,5] & Conv2D \\
        128 & [1,2] & [3,5] & Conv2D \\
        128 & [2,1] & [3,5] & Conv2D \\
        128 & - & - & Dense \\
        64 & - & - & Dense \\
        32 & - & - & Dense \\
        1 & - & - & Dense \\
        \hline
        \end{tabular}
        }
    \end{table}    

    The objective function for the generator is similar to Pix2Pix but instead of using the $L1$ loss, we introduce a dot product loss, in particular:
    \begin{equation}
        L_{dot}=flat(D(G(x,z))).flat(y) \label{eq5}
    \end{equation}
    
    The flat() function in \eqref{eq5} reshapes the image tensors to a vector in order to calculate the dot product. It has been found previously that such a loss function is especially useful in preserving the style of an image \cite{jing2019neural}, and is widely used in neural style transfer algorithms. For the case of homoglyph generation, we are trying to preserve the style of the target image, $y$, but yet would like additional variations so that a model trained with these augmented data would be less biased toward what any particular font can render. 
    
    The generator objective function is thus as follows:
    \begin{equation}
        G^* = min_G log(D(x,y)) + log(1-D(G(x,z)))+L_{dot} \label{eq6}
    \end{equation}
    
    For the discriminator, we find that removing the network's dependency on $x$ as a condition gives better results. Thus the objective function for the discriminator is as follows:
    \begin{equation}
        D^* = max_D log(D(y)) + log(1-D(G(x,z))) \label{eq7}
    \end{equation}
    
    A batch size of 64 was used and the network was trained over 25k steps.

    We also tried a CycleGAN approach. However, the images produced were not as varied as the above approach, which is based on Pix2Pix.

    \subsection{Homoglyph Identifier (HI)}
    The HI works by first encoding the input homoglyph image into a low dimensional vector. This is done through the use of a separately trained CNN, called the Encoder. We then apply the Encoder to a list of domains (referred to as the checking list), that we would like to protect against homoglyph attacks, to get reference vectors for each domain\footnote{The checking list that we use in this paper is: [google.com, linkedin.com, yahoo.com, wikipedia.org, apple.com, instagram.com, facebook.com, microsoft.com, twitter.com, youtube.com]}. Next, we calculate the Euclidean distance between the suspect homoglyph and each of the reference vectors. If the Euclidean distance is less than a threshold ($T$), we classify the suspect homoglyph as a homoglyph and we identify the domain that it was trying to mimic as the one with the smallest Euclidean distance.

    \begin{figure}[!htp]
        \centering
        \includegraphics[width=6cm]{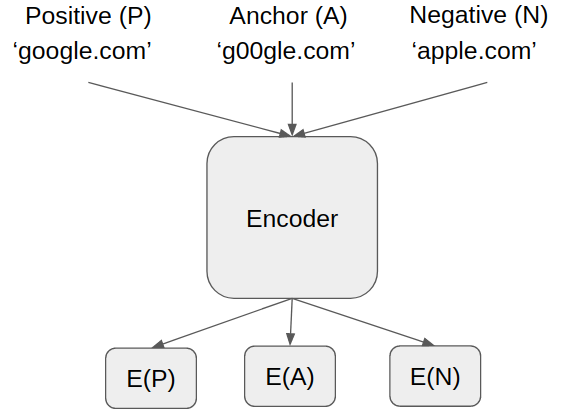}
        \caption{Encoder applied on Anchor, Positive \& Negative}
        \label{fig:encoder_sample}
    \end{figure}  
    
    We employ the triplet loss to train this network. Figure \ref{fig:encoder_sample} illustrates the triplet loss. In our formulation, we use the outputs of PhishGAN as the anchor. The positive sample is the example the anchor was trying to mimic in the checking list while the negative sample is sampled randomly from the checking list, with probabilities inversely proportional to the encoding's Euclidean distance from the anchor's encoding. Thus, an item in the checking list closer to the anchor's encoding ($E(A)$) in terms of Euclidean distance would have a higher probability of being sampled. It is important to note that only the output of PhishGAN is used to train the Encoder; no hand crafted dataset is used.

    \small
    \begin{equation}
        L_{triplet} = max(||E(A)-E(P)||^2  - ||E(A)-E(N)||^2 + M, 0.0) \label{eq8}
    \end{equation}
    \normalsize
    
    The Encoder is then trained via the loss function \eqref{eq8} \cite{schroff2015facenet}. It is clear from the equation that minimising $L_{triplet}$ is equivalent to ensuring that the distance between $E(A)$ and $E(P)$ is at least smaller than the distance between $E(A)$ and $E(N)$ with a margin of $M$. For the sake of brevity, we will not delve into the details of how this is an improvement over the pair loss in Siamese Networks as they are well documented in the literature \cite{hermans2017defense} \cite{triple_loss_v_contrastive_loss}.
    
    In our experiments, we set M to 1 arbitrarily, thus the threshold, $T$, was also set as 1. The network used for the encoder is also a CNN network identical to the one used by PhishGAN's discriminator, except that 3 dense layers of 128 neurons were used instead of the 4 stated in Table \ref{tab:discriminator_architecture}. The final activation of this CNN network is a linear one, which we subsequently activate via an $L2$ normalisation function. This ensures that the $L2$ norm of the encodings are always 1. Instance normalisation was used between network layers in this network.
    
    To analyse PhishGAN's impact on HI, we paint the following scenarios:
    \begin{itemize}
        \item Scenario 1: Given a checking list of 10 domains, we train a HI to identify homoglyphs of these domains and showcase the results on a testing dataset created via dnstwist\footnote{\url{https://github.com/elceef/dnstwist}} \footnote{Note: For all 3 scenarios, we constrained PhishGAN to only generate characters that Arial supports by constraining $x$ to be rendered in Arial. This is for fair evaluation of HI and PhishGAN, as glyphs produced by dnstwist are guaranteed renderable by either Arial or Times only.}
        \item Scenario 2: Next, we add an additional unseen domain, "covid19info.live" to the checking list. This is to observe the changes in performance due to the inclusion of a new domain that is likely to be used by hackers for phishing during the coronavirus pandemic in 2020.
        \item Scenario 3: Finally we train the HI again, using PhishGAN, with "covid19info.live" added to the checking list and observe the performance.
    \end{itemize}
    
\section{Results and Discussion} 
    \begin{figure}[!htp]
        \centering
        \includegraphics[width=9cm]{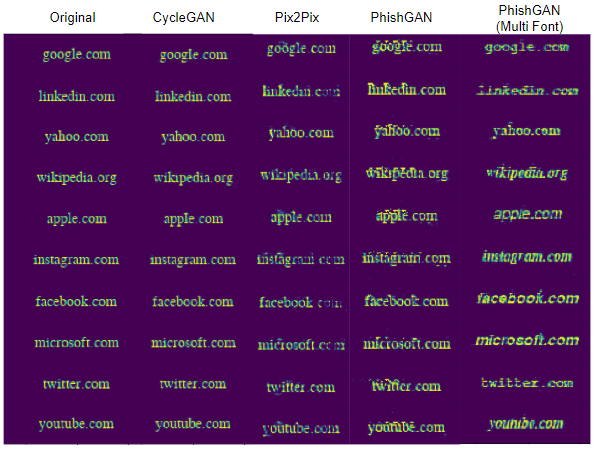}
        \caption{Sample outputs of PhishGAN compared to other methods}
        \label{fig:glyph_comparison}
    \end{figure}
    
    We first showcase sample outputs of PhishGAN for items in the checking list that this paper uses\footnote{As the checking list used in this paper was randomly generated based on popular domains the team surfs, we note that "yahoo.com", "wikipedia.org" and "microsoft.com" were not part of the data used to train PhishGAN} and compare them with the current SOTA. As can be seen in Figure \ref{fig:glyph_comparison}, PhishGAN, although similar in structure to Pix2Pix, is able to provide much more variations for training of homoglyph networks. The inability of CycleGAN to work shows that the variation between the $x$ and $y$ domains are very small and it may be difficult for an algorithm to automatically pick out these subtle differences. Pix2Pix was able to show some variation as can be seen from the small number of glyphs added to certain strings like "google.com", "apple.com", etc. It is obvious that PhishGAN produces the most varied number of glyphs. For "yahoo.com", it is interesting that it was able to produce "yaHoo.com". It was also able to morph "microsoft.com" quite significantly to something visually similar to "mlcrosoft.com". Finally, we showcase PhishGAN's ability to morph multiple fonts. It is interesting to see the example of "google.com", which was morphed to "gooale.com", "youtube.com" being morphed to "younibe.com" and "linkedin.com" to "iinkeain.com".
    
    We next use PhishGAN to train the HI. As mentioned in the previous section, we show the performance for the 3 scenarios. We will use 2 evaluation metrics; first we determine its ability in detecting homoglyphs\footnote{It does not make sense to check the HI's ability to pick out non homoglyphs as the Euclidean distance would be 0. Thus, the accuracy metric was used to determine its ability to pick out homoglyphs.}. Next, we determine its ability to classify those homoglyphs that were detected to those in the checking list. It is important to note that the same model was used in both Scenario 1 and Scenario 2 while Scenario 3 pertains to retraining a new model, taking into account the additional domain in the checking list. All models were trained to convergence in terms of training loss (i.e. triplet loss).

    \begin{table}[!htbp]
    \caption{Accuracy \& F1-Score of various Experiments and Scenarios}
    \label{tab:accuracy_n_f1_score}
    \centerline{
        \begin{tabular}{|p{1.3cm}|p{1.7cm}|p{1.2cm}|p{1.2cm}|p{1.2cm}|}
        \hline
        \textbf{ } & \textbf{Experiment} & \textbf{Scenario 1} & \textbf{Scenario 2} & \textbf{Scenario 3}\\
        \hline
        Accuracy & 1 & 0.83 & 0.76 & 0.89 \\
                 & 2 & 0.80 & 0.77 & 0.86 \\
                 & 3 & 0.81 & 0.78 & 0.85 \\
        \hline
        Mean Acc & & 0.81 & 0.76 & 0.86 \\
        \hline
        F1-Score & 1 & 0.77 & 0.76 & 0.87 \\
                 & 2 & 0.78 & 0.85 & 0.89 \\
                 & 3 & 0.81 & 0.87 & 0.84 \\
        \hline
        Mean F1 & & 0.78 & 0.83 & 0.86 \\
        \hline
        \end{tabular}
        }
    \end{table}        

    As can be seen in Table \ref{tab:accuracy_n_f1_score}, there is a degradation in accuracy when moving from scenario 1 to scenario 2. This could be due to the fact that the model was not trained on "covid19info.live" in the checking list. Thus, it is less aware of the type of glyphs one may expect from such a domain. There is largely an increase in the F1 score because out of those that were identified as homoglyphs, more were classified correctly on average. Next, we retrained the model in scenario 3 and we see that we are able to regain the accuracy performance and also increase F1 score. This could be due to the larger training dataset as the model now also sees glyphs from one other example, which may help in detecting glyphs of another.

\begin{figure}[!htp]
        \centering
        \includegraphics[width=8cm]{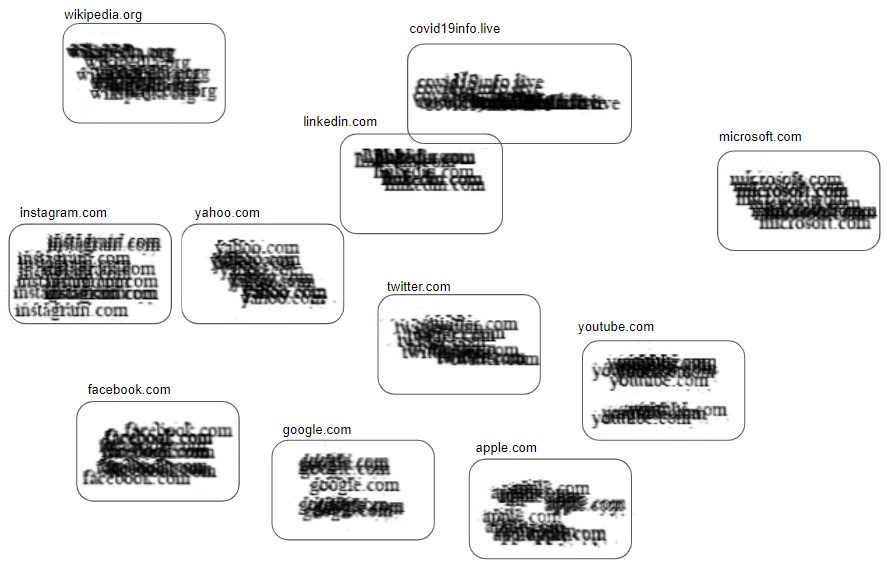}
        \caption{Clustering of domains generated by PhishGAN}
        \label{fig:clustered_domains}
    \end{figure}

    Finally, we showcase HI's ability to produce meaningful encodings of the input text based images. As can be seen in Figure \ref{fig:clustered_domains}, the Encoder is able to encode the different homoglyph images, produced by PhishGAN, into clearly discernible clusters, with each cluster corresponding to a particular domain name as shown in Figure \ref{fig:clustered_domains}. The variation within each cluster also showcases PhishGAN's ability to generate a variety of homoglyphs for each string. This shows that PhishGAN is not suffering from mode collapse \footnote{\url{https://developers.google.com/machine-learning/gan/problems}}, a widely known problem for GANs. On this note, it should be added that the original Pix2Pix architecture did not include the addition of noise tensors as the authors\footnote{\url{https://github.com/junyanz/pytorch-CycleGAN-and-pix2pix/issues/152}} find that they do not cause significant variations in the output, given a particular input condition, $x$. In our case, however, the additional noise tensors were important as they allowed varied outputs as evidenced in Figure \ref{fig:clustered_domains}, where clusters aren't just a single point. Finally, Figure \ref{fig:clustered_domains} also shows why there could be misclassifications. Looking at the "linkedin.com" and "covid19info.live" clusters, we observe that the distance could be quite near, indicating the possibility of classification errors.
    
\section{Conclusion}
    We have shown how PhishGAN outperforms current SOTA in its ability to generate homoglyphs. This was achieved by making practical modifications to the Pix2Pix architecture:
    \begin{enumerate}
        \item Replacing $L1$ loss with $L_{dot}$.
        \item Inclusion of noise tensors in the generator.
        \item Removing condition input, $x$, from the discriminator.
        \item Using instance normalisation for the discriminator.
    \end{enumerate}
    We also show that we are able to get reasonable performances using just PhishGAN generated data to train a HI. We then showcase how adding an additional domain into the checking list may degrade HI's accuracy performance and how PhishGAN can be used to retrain the HI to regain its accuracy and classification performance. The HI's ability to encode the text based images into discernible clusters was also verified by applying t-SNE on the encodings, validating that the triplet loss is indeed an appropriate loss for this problem setup. Finally, the variations within each cluster are also proof that PhishGAN doesn't suffer from mode collapse problems that plague GANs in general. The work here has significant impact for future research into homoglyph detection as it shows that deep neural networks can also be used to generate homoglyphs which can in turn be used to train homoglyph detection models and update them on the fly as new threats emerge. We believe the work done here provides a significant alternative to handcrafting homoglyph datasets and can contribute significantly to the prevention of homoglyph phishing attacks. To the best of our knowledge, this is also the first piece of work that applies GANs to text based images, opening up a new research direction for the application of GANs. One limitation is that PhishGAN currently utilizes only the English language in it's generation and detection. Hence, future work could extend PhishGAN to other languages such as Chinese \& Korean which have vastly varying characters. Furthermore, relaxing the need for paired training data to train the GAN should be explored to reduce the effort required to create a training dataset for the GAN.

\bibliographystyle{unsrtnat}
\bibliography{main}

\begin{thebibliography}{10}
\providecommand{\natexlab}[1]{#1}
\providecommand{\url}[1]{\texttt{#1}}
\expandafter\ifx\csname urlstyle\endcsname\relax
  \providecommand{\doi}[1]{doi: #1}\else
  \providecommand{\doi}{doi: \begingroup \urlstyle{rm}\Url}\fi

\bibitem[Dhamija et~al.(2006)Dhamija, Tygar, and Hearst]{dhamija2006phishing}
Rachna Dhamija, J~Doug Tygar, and Marti Hearst.
\newblock Why phishing works.
\newblock In \emph{Proceedings of the SIGCHI conference on Human Factors in
  computing systems}, pages 581--590, 2006.

\bibitem[Suzuki et~al.(2019)Suzuki, Chiba, Yoneya, Mori, and Goto]{shamfinder}
Hiroaki Suzuki, Daiki Chiba, Yoshiro Yoneya, Tatsuya Mori, and Shigeki Goto.
\newblock Shamfinder: An automated framework for detecting idn homographs.
\newblock \emph{ACM Internet Measurement Conference}, 2019.

\bibitem[Woodbridge et~al.(2018)Woodbridge, Anderson, Ahuja, and
  Grant]{woodbridge2018detecting}
Jonathan Woodbridge, Hyrum~S Anderson, Anjum Ahuja, and Daniel Grant.
\newblock Detecting homoglyph attacks with a siamese neural network.
\newblock In \emph{2018 IEEE Security and Privacy Workshops (SPW)}, pages
  22--28. IEEE, 2018.

\bibitem[Isola et~al.(2017)Isola, Zhu, Zhou, and Efros]{isola2017image}
Phillip Isola, Jun-Yan Zhu, Tinghui Zhou, and Alexei~A Efros.
\newblock Image-to-image translation with conditional adversarial networks.
\newblock In \emph{Proceedings of the IEEE conference on computer vision and
  pattern recognition}, pages 1125--1134, 2017.

\bibitem[He et~al.(2015)He, Zhang, Ren, and Sun]{resnet_ref}
Kaiming He, Xiangyu Zhang, Shaoqing Ren, and Jian Sun.
\newblock Deep residual learning for image recognition.
\newblock \emph{CoRR}, abs/1512.03385, 2015.
\newblock URL \url{http://arxiv.org/abs/1512.03385}.

\bibitem[Zhu et~al.(2017)Zhu, Park, Isola, and Efros]{zhu2017unpaired}
Jun-Yan Zhu, Taesung Park, Phillip Isola, and Alexei~A Efros.
\newblock Unpaired image-to-image translation using cycle-consistent
  adversarial networks.
\newblock In \emph{Proceedings of the IEEE international conference on computer
  vision}, pages 2223--2232, 2017.

\bibitem[Wu et~al.(2017)Wu, Manmatha, Smola, and
  Kr{\"{a}}henb{\"{u}}hl]{triple_loss_v_contrastive_loss}
Chao{-}Yuan Wu, R.~Manmatha, Alexander~J. Smola, and Philipp
  Kr{\"{a}}henb{\"{u}}hl.
\newblock Sampling matters in deep embedding learning.
\newblock \emph{CoRR}, abs/1706.07567, 2017.
\newblock URL \url{http://arxiv.org/abs/1706.07567}.

\bibitem[Jing et~al.(2019)Jing, Yang, Feng, Ye, Yu, and Song]{jing2019neural}
Yongcheng Jing, Yezhou Yang, Zunlei Feng, Jingwen Ye, Yizhou Yu, and Mingli
  Song.
\newblock Neural style transfer: A review.
\newblock \emph{IEEE transactions on visualization and computer graphics},
  2019.

\bibitem[Schroff et~al.(2015)Schroff, Kalenichenko, and
  Philbin]{schroff2015facenet}
Florian Schroff, Dmitry Kalenichenko, and James Philbin.
\newblock Facenet: A unified embedding for face recognition and clustering.
\newblock In \emph{Proceedings of the IEEE conference on computer vision and
  pattern recognition}, pages 815--823, 2015.

\bibitem[Hermans et~al.(2017)Hermans, Beyer, and Leibe]{hermans2017defense}
Alexander Hermans, Lucas Beyer, and Bastian Leibe.
\newblock In defense of the triplet loss for person re-identification.
\newblock \emph{arXiv preprint arXiv:1703.07737}, 2017.

\end{thebibliography}

\end{document}